\documentclass[runningheads]{llncs}
\usepackage{colortbl}
\usepackage[table]{xcolor}
\usepackage{multirow}
\usepackage{amsmath,amssymb}

\usepackage[T1]{fontenc}
\usepackage{graphicx}
\begin{document}
\title{Towards Globally Predictable $k$-Space Interpolation: A White-box Transformer Approach}
\titlerunning{Towards Globally Predictable $k$-Space Interpolation}
%
\author{Chen Luo\inst{1,\dagger} \and Qiyu Jin\inst{1,\dagger} \and Taofeng Xie\inst{2} \and Xuemei Wang\inst{3} \and Huayu Wang\inst{4} \and Congcong Liu\inst{4} \and Liming Tang\inst{5} \and Guoqing Chen\inst{1}  \and  Zhuo-Xu Cui\inst{4,6,*} \and Dong Liang\inst{4,6,*}}
%

\authorrunning{C. Luo et al.}
%
\institute{School of Mathematical Sciences, Inner Mongolia University, China \and
College of Computer and Information, lnner Mongolia Medical University, China \and
Inner Mongolia Medical University Affiliated Hospital, China \and
Shenzhen Institutes of Advanced Technology, Chinese Academy of Sciences, China \and
School of Mathematics and Statistics, Hubei Minzu University, China \and
Key Laboratory of Biomedical Imaging Science and System, Chinese Academy of Sciences, China \\
\email{zx.cui@siat.ac.cn, dong.liang@siat.ac.cn} }

\maketitle              
\renewcommand{\thefootnote}{}%
\footnotetext{\textsuperscript{\dag} These authors contributed equally to this work.}
\footnotetext{\textsuperscript{*} Corresponding authors.}

\begin{abstract}
Interpolating missing data in \( k \)-space is essential for accelerating imaging. However, existing methods, including convolutional neural network-based deep learning, primarily exploit local predictability while overlooking the inherent global dependencies in \( k \)-space. Recently, Transformers have demonstrated remarkable success in natural language processing and image analysis due to their ability to capture long-range dependencies. This inspires the use of Transformers for \( k \)-space interpolation to better exploit its global structure. However, their lack of interpretability raises concerns regarding the reliability of interpolated data. To address this limitation, we propose GPI-WT, a white-box Transformer framework based on \textit{Globally Predictable Interpolation (GPI)} for \( k \)-space. Specifically, we formulate GPI from the perspective of annihilation as a novel \( k \)-space structured low-rank (SLR) model. The global annihilation filters in the SLR model are treated as learnable parameters, and the subgradients of the SLR model naturally induce a learnable attention mechanism. By unfolding the subgradient-based optimization algorithm of SLR into a cascaded network, we construct the first white-box Transformer specifically designed for accelerated MRI. Experimental results demonstrate that the proposed method significantly outperforms state-of-the-art approaches in \( k \)-space interpolation accuracy while providing superior interpretability.

\end{abstract}
\section{Introduction}
Magnetic resonance imaging (MRI) is a cornerstone clinical imaging modality, widely recognized for its non-invasive nature, absence of ionizing radiation, and exceptional soft tissue contrast. However, the inherently prolonged acquisition time imposes significant practical limitations, often requiring the acquisition of only partial \( k \)-space data~\cite{Hutchinson1988Fast,Lustig2007Sparse}. Consequently, advanced reconstruction algorithms are employed to recover high-quality images from these incomplete measurements~\cite{liang1992constrained,Aggarwal2019MoDL}. A central challenge in MRI reconstruction is the accurate prediction of missing \( k \)-space data.

The efficacy of \( k \)-space interpolation methods fundamentally relies on the assumption that missing \( k \)-space data can be reliably predicted~\cite{Griswold2002Generalized}. Traditional approaches, such as those based on image sparsity~\cite{Lustig2007Sparse,Lustig2008Compressed,Uecker2014ESPIRiT}, phase smoothness~\cite{Jacob2020Structured,Shin2014SAKE,Lee2016ALOHA,haldar2013low,Haldar2016PLORAKS}, and coil sensitivity smoothness~\cite{Pruessmann1999Sense,Pruessmann2001Advances}, have enabled the estimation of missing \( k \)-space data from local neighboring information~\cite{Smith1986Application,Griswold2002Generalized,Lustig2010Spirit}. However, these assumptions often do not hold rigorously in practical scenarios. When these assumptions are violated, the predictability of \( k \)-space data transitions from a localized interpolation problem to a more complex global dependency, leading to significant errors and degraded image quality.

Recent advancements have explored deep learning (DL)-based methodologies for \( k \)-space interpolation~\cite{Han2020kSpaceDL,DU2021Adaptive,Zhao2022KspaceTrans,Cui2023KUNN,Eo2018KIKI}. Notably, studies such as~\cite{Pramanik2020DeepSLR,Kim2019Loraki,Chen2023Matrix} have replaced the linear interpolation in traditional methods with nonlinear convolutional neural networks (CNNs), leveraging large-scale datasets to enhance interpolation accuracy. Despite their effectiveness in capturing local structures, CNNs inherently possess a limited receptive field, making them insufficient for modeling long-range dependencies in \( k \)-space. This limitation highlights the need for alternative approaches capable of capturing global relationships within \( k \)-space.

Transformer-based models~\cite{vaswani2017attention} have shown remarkable success in capturing long-range dependencies through self-attention mechanisms and fully connected layers, making them highly promising for \( k \)-space interpolation. However, existing Transformer-based approaches~\cite{Liu2021SwinTransformer,Zhao2022KspaceTrans,WU2023DeepTrans,Guo2024ReconFormer} typically function as black-box models, lacking theoretical interpretability, which raises concerns regarding their reliability in \( k \)-space interpolation tasks. A significant advancement by Yu et al.~\cite{Yu2023WhiteBox} introduced a white-box Transformer for image classification, optimizing sparse rate reduction and providing a theoretical foundation for multi-head self-attention in terms of encoding efficiency. This work also links self-attention mechanisms to iterative optimization algorithms. Although the concept of sparse rate reduction is not directly applicable to \( k \)-space interpolation, this insight inspires the development of a novel, interpretable white-box Transformer tailored to the global interpolation properties inherent in \( k \)-space.

In this paper, we introduce a novel \( k \)-space structured low-rank (SLR) model, formulated from the perspective of globally predictable interpolation (GPI) in \( k \)-space through annihilation. Within this framework, the annihilation filters are defined as learnable parameters, and the subgradients of the SLR model naturally induce a learnable attention mechanism. By unfolding the subgradient-based optimization process for SLR-based \( k \)-space interpolation into a cascaded network, we develop a white-box Transformer specifically designed for \( k \)-space interpolation.

The main contributions of this paper are as follows: (1)~\textit{Interpretable white-box Transformer for MRI reconstruction}: We introduce the first theoretically grounded white-box Transformer model specifically designed for MRI reconstruction.
(2)~\textit{Effective modeling of global dependencies}: Unlike conventional CNN-based \( k \)-space interpolation methods, the proposed Transformer model effectively captures long-range dependencies within \( k \)-space, enabling more accurate interpolation of missing data. (3)~\textit{Comprehensive experimental validation}: The proposed approach is trained and evaluated on MRI datasets, and benchmarked against state-of-the-art methods. Experimental results demonstrate that the proposed GPI-WT consistently outperforms competing approaches across various undersampling patterns, achieving superior performance in both qualitative and quantitative assessments.

\section{Methods}\label{sec: method}
\subsection{Globel White-box Transformer Reconstruction Model}
For parallel MRI, the forward model of \( k \)-space measurements can be mathematically represented as follows:
\begin{equation}\label{eq:1}
\mathbf{y} = M_{\Omega}\mathbf{k} + \mathbf{n},
\end{equation}
where \( \mathbf{k} = [\mathbf{k}_1, \ldots, \mathbf{k}_{N_c}] \) and \( \mathbf{y} \in \mathbb{C}^{N_1 \times N_2 \times N_c} \) denote the multi-coil fully sampled \( k \)-space data and the undersampled \( k \)-space data, respectively, with \( N_c \geq 1 \) coils. \( \mathbf{n} \) represents the noise. The sampling pattern \( M_{\Omega} \in \mathbb{C}^{N_1 \times N_2 \times N_c} \) is set to 1 at the sampled positions and 0 otherwise. We assume that the missing \( k \)-space data in \( \mathbf{y} \) can be reliably predicted, which forms the basis for accurately reconstructing \( \mathbf{k} \) from \( \mathbf{y} \).

The SLR model, conceptualized from the perspective of globally predictable interpolation in \( k \)-space through annihilation, can be formulated as a summation \( \sum_{h=1}^H\|\mathcal{H}(\mathbf{k},d)\mathbf{s}_h\|_F^2 \), where \( \mathcal{H} \) represents the Hankelization operation, \( d \) is the window size used for sliding over the data, and $\mathbf{s}_h$ are the annihilation filters~\cite{ye2018deep}. In this framework, we consider the globally predictable and interpolated annihilation dependencies in \( k \)-space, thus \( d \) is chosen to match the size of the \( k \)-space data. A well-established result is that \( \|\mathcal{H}(\mathbf{k},d)\mathbf{s}_h\|_F^2 = \|\mathbf{Q}_h\mathbf{k}\|_F^2 \), where \( \mathbf{Q}_h \) denotes the Hankel transformation of the annihilating filter \( \mathbf{s}_h \)~\cite{Pramanik2020DeepSLR}. By leveraging the equivalence between the Frobenius norm and the trace of a matrix \( \|\mathbf{X}\|_F^2 = \text{Tr}(\mathbf{X}^*\mathbf{X}) \), the SLR constraint \( \|\mathbf{Q}_h\mathbf{k}\|_F^2 \) can be reformulated as \( \text{Tr}\left[ \left(\mathbf{Q}_h \mathbf{k}\right)^*\left(\mathbf{Q}_h \mathbf{k}\right)\right] \).

To further refine the SLR model, we introduce a penalty function \( \rho_\gamma \) applied to \( \mathbf{Q}\mathbf{k} \), extending the SLR constraint to:

\begin{equation}\label{slr:1}
    R\left(\mathbf{k};\mathbf{Q}_{[H]}\right):=\sum_{h=1}^H\text{Tr}\left[ \rho_\gamma\left(\left(\mathbf{Q}_h \mathbf{k}\right)^*\left(\mathbf{Q}_h \mathbf{k}\right)\right)\right]
\end{equation}
where \( \rho_\gamma(\mathbf{X}) = \mathbf{U \text{Diag}(\rho_\gamma(\sigma_1),...,\rho_\gamma(\sigma_r))}\mathbf{V}^* \) and \( \sigma_i \) denotes the singular values of \( \mathbf{X} \). In this work, we adopt a sparse prompting function $$
\rho_\gamma(\mathbf{X}) := \mathbf{U \text{diag}(\ln(1+\gamma\sigma_1), \dots, \ln(1+\gamma\sigma_r))}\mathbf{V}^*.$$

In traditional methods, \( \mathbf{Q}_h \) is obtained by applying Hankelization to low-dimensional filters \( \mathbf{s}_h \), leveraging the local predictability of \( k \)-space. This approach involves a small number of parameters, which can typically be estimated from calibration data. However, in our method, \( \mathbf{s}_h \) is treated as a global annihilation filter with dimensions matching those of the \( k \)-space data \( \mathbf{k} \), resulting in a substantially larger parameter space. This increased complexity makes it difficult to accurately estimate \( \mathbf{s}_h \) using calibration data alone. To address this, we relax the Hankel structural constraint on \( \mathbf{Q}_h \) and define it as a set of learnable parameters, allowing it to be directly learned from large datasets.

On the other hand, the subgradient of the specially designed SLR model (\ref{slr:1}) can be approximated as the following structure:
\begin{equation}\begin{aligned}
\nabla_{\mathbf{k}}  R\left(\mathbf{k};\mathbf{Q}_{[H]}\right)
    &= \sum_{h=1}^H\nabla_{\mathbf{k}} \text{Tr}\left[ \ln\left(\mathbf{I}+\gamma\left(\mathbf{Q}_h \mathbf{k} \right)^* \left(\mathbf{Q}_h \mathbf{k}\right)\right)\right] \\
    &= \sum_{h=1}^H\nabla_{\mathbf{k}} \ln\text{Det}\left[ \left(\mathbf{I}+\gamma\left(\mathbf{Q}_h \mathbf{k} \right)^* \left(\mathbf{Q}_h \mathbf{k}\right)\right)\right] \\
    &=\gamma\sum_{h=1}^H \mathbf{Q}_h^*\mathbf{Q}_h \mathbf{k} \left(\mathbf{I}+\gamma\left(\mathbf{Q}_h \mathbf{k} \right)^* \left(\mathbf{Q}_h \mathbf{k}\right)\right)^{-1} \\
   &\approx \gamma\mathbf{k} - \gamma^2\sum_{h=1}^H\mathbf{Q}_{h}^{*}\mathbf{Q}_{h} \mathbf{k} \operatorname{softmax}\left(\left(\mathbf{Q}_{h} \mathbf{k} \right)^*\mathbf{Q}_{h} \mathbf{k}\right), 
\end{aligned}\end{equation} 
where the last approximate equality follows from reference \cite{Yu2023WhiteBox}. Based on the above analysis, the subgradient of (\ref{slr:1}) can fundamentally be interpreted as a multi-head subspace self-attention (MSSA):
\begin{equation}\label{eq:6}
\text{MSSA}(\mathbf{k}|\mathbf{Q}_{[H]}) := \gamma^2[\mathbf{Q}^*_1,\ldots,\mathbf{Q}^*_H]\begin{bmatrix}
 \text{SSA}(\mathbf{k}|\mathbf{Q}_{1})\\
 \vdots  \\
 \text{SSA}(\mathbf{k}|\mathbf{Q}_{H})
\end{bmatrix},
\end{equation}
where subspace self-attention (SSA) is defined as
\begin{equation}\label{eq:7}
 \text{SSA}(\mathbf{k}|\mathbf{Q}_{h}) := \mathbf{Q}_{h} \mathbf{k} \, \text{softmax} \left(\left( \mathbf{Q}_{h} \mathbf{k} \right)^*\mathbf{Q}_{h} \mathbf{k}\right).
\end{equation}
Unlike conventional Transformers, the MSSA mechanism utilizes a single matrix to derive the Query, Key, and Value representations in the attention mechanism of a white-box Transformer, where \( Q = K = V = \mathbf{Q} \mathbf{k} \). Therefore, we derive an interpretable white-box attention mechanism through the SLR model (\ref{slr:1}).

\begin{figure}[t]
\includegraphics[width=\textwidth]{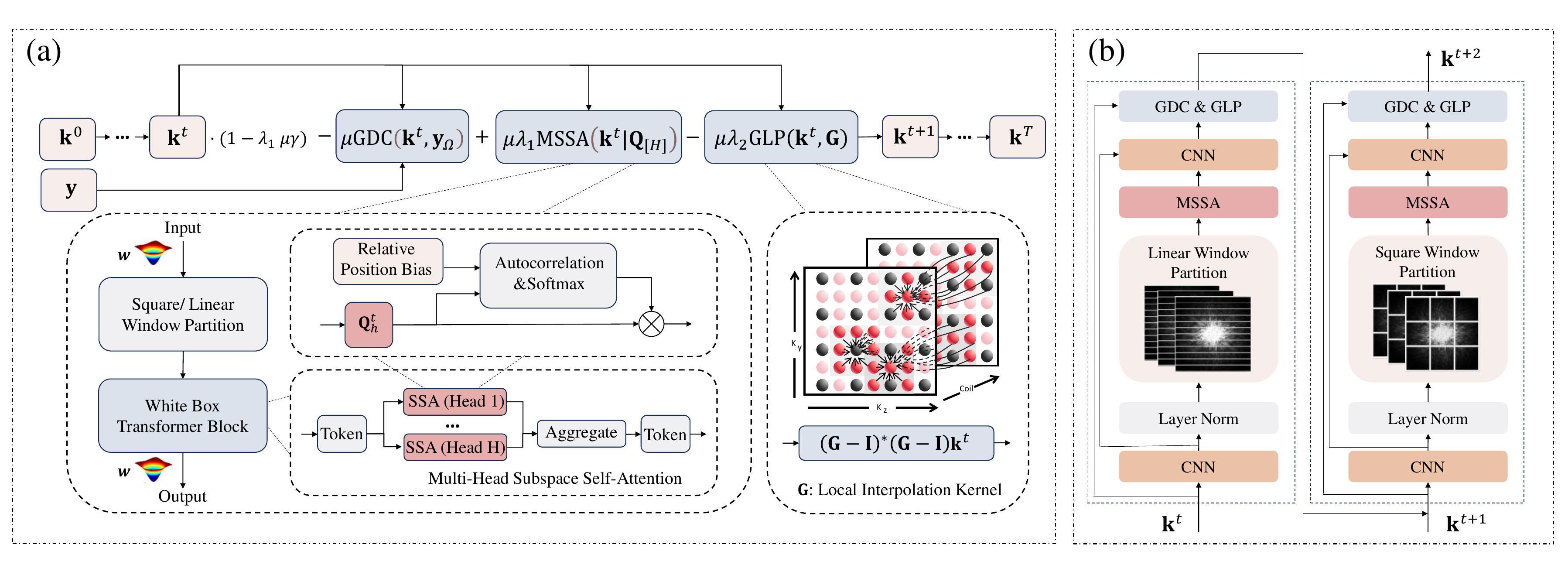}
\caption{(a) Schematic of the our proposed unfolded network. (b) Diagram of window partitions. Smoothing filter $w$ is applied to facilitate the computation in $k$-space. } \label{fig}
\end{figure}

Based on the white-box attention mechanism constructed above, by unfolding the subgradient-based optimization algorithm for the (\ref{slr:1})-regularized \( k \)-space interpolation model, we derive a cascaded white-box Transformer network specifically tailored for \( k \)-space interpolation. In particular, the (\ref{slr:1})-regularized \( k \)-space interpolation model can be formulated as follows:

\begin{equation}\label{eq:slr+sc}
\min_{\mathbf{k}} \frac{1}{2}\|M_{\Omega}\mathbf{k} - \mathbf{y}\|_2^2 + \lambda_1R\left(\mathbf{k};\mathbf{Q}_{[H]}\right)   + \lambda_2 \| (\mathbf{G}-\mathbf{I})\mathbf{k} \|_2^2,
\end{equation}
It is worth noting that, considering that the global predictability enforced by (\ref{slr:1}) may weaken local dependencies, we have retained a local predictability  term \( \| (\mathbf{G}-\mathbf{I})\mathbf{k} \|_2^2 \) in the above model. Here, \( \mathbf{G} \) represents the local linear interpolation kernel estimated by SPIRiT \cite{Lustig2010Spirit}. The parameters \( \lambda_1 \) and \( \lambda_2 \) are regularization coefficients. The gradient descent method for (\ref{eq:slr+sc}) yields a cascaded white-box Transformer, as detailed below:

\begin{equation}\label{eq:8}
    \mathbf{k}^{t+1} =(1-\lambda_1\mu\gamma) \mathbf{k}^{t} - \mu \text{GDC}(\mathbf{k}^{t}, \mathbf{y}) + \mu\lambda_1 \text{MSSA}(\mathbf{k}^{t}| \mathbf{Q}_{[H]}) 
    -\mu\lambda_2 \text{GLP}(\mathbf{k}^{t}, \mathbf{G}),
\end{equation}
where \( \mu \) denotes the step size, $\text{GDC}$ is the gradient of the data consistency term, i.e., 
$
\text{GDC}(\mathbf{k}, \mathbf{y}) =  M_{\Omega}^* M_{\Omega} \mathbf{k} -  M_{\Omega}^*\mathbf{y},
$
and $\text{GLP}$ is the gradient of the local predictability term, i.e.,
$
\text{GLP}(\mathbf{k}, \mathbf{G}) = (\mathbf{G}-\mathbf{I})^{*}(\mathbf{G}-\mathbf{I})\mathbf{k}.
$
The specific structural framework of this approach is illustrated in Fig.~\ref{fig}(a).

\subsection{Network Architecture} 
\subsubsection{Square and linear window partiton}
In visual Transformer-based methods \cite{Dosovitskiy2021ViT,Yu2023WhiteBox}, images are divided into patches, leading to quadratic computational growth as image size increases. To address this, we adopt Swin Transformer's \cite{Liu2021SwinTransformer} window partition, treating each pixel as a patch within fixed-size non-overlapping windows. This allows our MSSA to maintain linear computational scaling.

Unlike the image domain, where neighbouring pixels are highly correlated, \( k \)-space exhibits long-range dependencies between symmetric signals. Standard self-attention with square windows limits cross-window interactions, restricting distant information capture. To overcome this limitation, we propose a novel linear window strategy that preserves efficiency while enhancing attention to distant signals by using non-overlapping lines as windows. Linear and square partitions alternate in successive GPI-WT iterations, improving the network’s nonlinear representation and reconstruction performance. Further details are illustrated in Fig.~\ref{fig}(b).  

\subsubsection{Relative position bias}
As the Transformer architecture is inherently order-agnostic and relies on self-attention mechanisms, it is essential to explicitly incorporate positional information through positional embeddings. In our approach, we use learnable relative position bias~\cite{Liu2021SwinTransformer} when computing self-attention, which not only reduces the number of trainable parameters but also enhances the performance of our model in square and linear window partitions.
Relative position bias $\mathbf{B}$ is incorporated into each head of MSSA, thus the SSA module can ultimately be modified to:
\begin{equation}\label{eq:9}
 \text{SSA}(\mathbf{k}|\mathbf{Q}_{h}) = \mathbf{Q}_{h} \mathbf{k} \text{softmax} \left(\left( \mathbf{Q}_{h} \mathbf{k} \right)^*\mathbf{Q}_{h} \mathbf{k} + \mathbf{B} \right).
\end{equation}

\section{Data and Experiments}
The raw knee data \footnote{\url{https://fastmri.org/}} was acquired from a 3T Siemens scanner. Data acquisition used a 15 channel knee coil array and conventional Cartesian 2D TSE protocol. We randomly selected 31 individuals (840 slices in total) as training data and 3 individuals (96 slices in total) as test data. The image size is cropped to $320 \times 300$.

The model unfolded $10$ iterations empirically with $4 \times 4$ square window and $6$ head in self-attention. We used ADAM optimizer \cite{kingma2014adam} starting with learning rate $lr=0.0001$, which decayed exponentially at a rate of $0.99$. Experiments were on Nvidia Tesla A6000 GPU. 
Three metrics were used to evaluate the results quantitatively, including normalized mean square error (NMSE), the peak signal-to-noise ratio (PSNR), and the structural similarity index (SSIM) \cite{Wang2004SSIM}.

\section{Results and Discussion}

\begin{table}[t]
\caption{Quantitative comparison for various methods on knee dataset with different mask patterns $M_{\Omega}$ and AF. $\mathcal{R}$: random mask. $\mathcal{U}$: uniform mask. The optimal values are denoted in \textbf{bold} and the proposed method is highlighted with gray background.}\label{tab: knee_random}
\centering 
{\fontsize{8}{10}\selectfont 
\begin{tabular}{l|c|cc|cc|cc}
  \hline
  \multirow{2}{*}{Method} & \multirow{2}{*}{$M_{\Omega}$} & \multicolumn{2}{c|}{NMSE $(\%)\downarrow$ } &\multicolumn{2}{c|}{PSNR $\uparrow$} & \multicolumn{2}{c}{SSIM $(\%)\uparrow$} \\
  \cline{3-8}
  \multirow{2}{*}{} & \multirow{2}{*}{} & AF=4 & AF=6 & AF=4 & AF=6 & AF=4 & AF=6 \\
  \hline  
   SPIRiT  &   \multirow{6}{*}{$\mathcal{R}$}   & 1.59$\pm$1.20 & 1.71$\pm$1.00 & 29.63$\pm$3.15 & 28.95$\pm$2.50 & 73.33$\pm$10.28 & 73.01$\pm$8.50 \\
   KNet   &   \multirow{6}{*}{}    & 0.84$\pm$0.30 & 1.38$\pm$0.68 & 31.73$\pm$1.44 & 29.75$\pm$1.78 & 85.28$\pm$3.08 & 80.91$\pm$3.95 \\ 
   Swin   &   \multirow{6}{*}{}    & 0.74$\pm$0.38 & 1.17$\pm$0.57 & 32.46$\pm$1.84 & 30.40$\pm$1.70 & 85.86$\pm$4.10 & 81.29$\pm$4.23 \\ 
   DSLR    &   \multirow{6}{*}{}    & 0.76$\pm$0.28 & 1.23$\pm$0.50 & 32.19$\pm$1.64 & 30.13$\pm$1.77 & 85.84$\pm$3.28 & 81.02$\pm$4.02 \\ 
   GPI-CNN  &   \multirow{6}{*}{}    & 0.57$\pm$0.29 & 0.96$\pm$0.38 & 33.48$\pm$1.69 & 31.18$\pm$1.67 & 87.42$\pm$2.87 & 83.41$\pm$3.27 \\ 
   \rowcolor{gray!20} \bfseries GPI-WT  &   \multirow{6}{*}{} & \bfseries 0.48$\pm$0.18 & \bfseries 0.81$\pm$0.33 & \bfseries 34.13$\pm$1.64 & \bfseries 31.92$\pm$1.65 & \bfseries 88.94$\pm$3.24 & \bfseries 84.93$\pm$3.84 \\ 
  \hline  
   SPIRiT &   \multirow{6}{*}{$\mathcal{U}$}   & 1.53$\pm$1.00 & 2.23$\pm$1.02 & 29.65$\pm$3.02 & 27.58$\pm$2.08 & 74.88$\pm$9.39 & 71.99$\pm$7.28 \\
   KNet    &   \multirow{6}{*}{}    & 1.13$\pm$0.55 & 2.32$\pm$1.04 & 30.60$\pm$1.70 & 27.52$\pm$1.83 & 84.35$\pm$3.14 & 76.88$\pm$4.03 \\ 
   Swin    &   \multirow{6}{*}{}    & 0.87$\pm$0.44 & 2.08$\pm$1.20 & 31.72$\pm$1.85 & 28.06$\pm$1.91 & 85.65$\pm$4.16 & 77.75$\pm$4.55 \\ 
   DSLR    &   \multirow{6}{*}{}    & 1.09$\pm$0.52 & 2.17$\pm$0.87 & 30.74$\pm$1.78 & 27.72$\pm$1.68 & 84.54$\pm$3.25 & 76.70$\pm$4.25 \\ 
   GPI-CNN  &   \multirow{6}{*}{}    & 0.69$\pm$0.46 & 1.36$\pm$0.59 & 32.79$\pm$1.86 & 29.77$\pm$1.73 & 87.31$\pm$3.05 & 80.98$\pm$3.24 \\ 
   \rowcolor{gray!20} \bfseries GPI-WT  &   \multirow{6}{*}{}  & \bfseries 0.55$\pm$0.25 & \bfseries 1.10$\pm$0.41 & \bfseries 33.64$\pm$1.78 & \bfseries 30.59$\pm$1.39 & \bfseries 88.79$\pm$3.30 & \bfseries 82.97$\pm$3.48 \\ 
  \hline  
\end{tabular}}
\end{table}

\begin{figure}
\includegraphics[width=\textwidth]{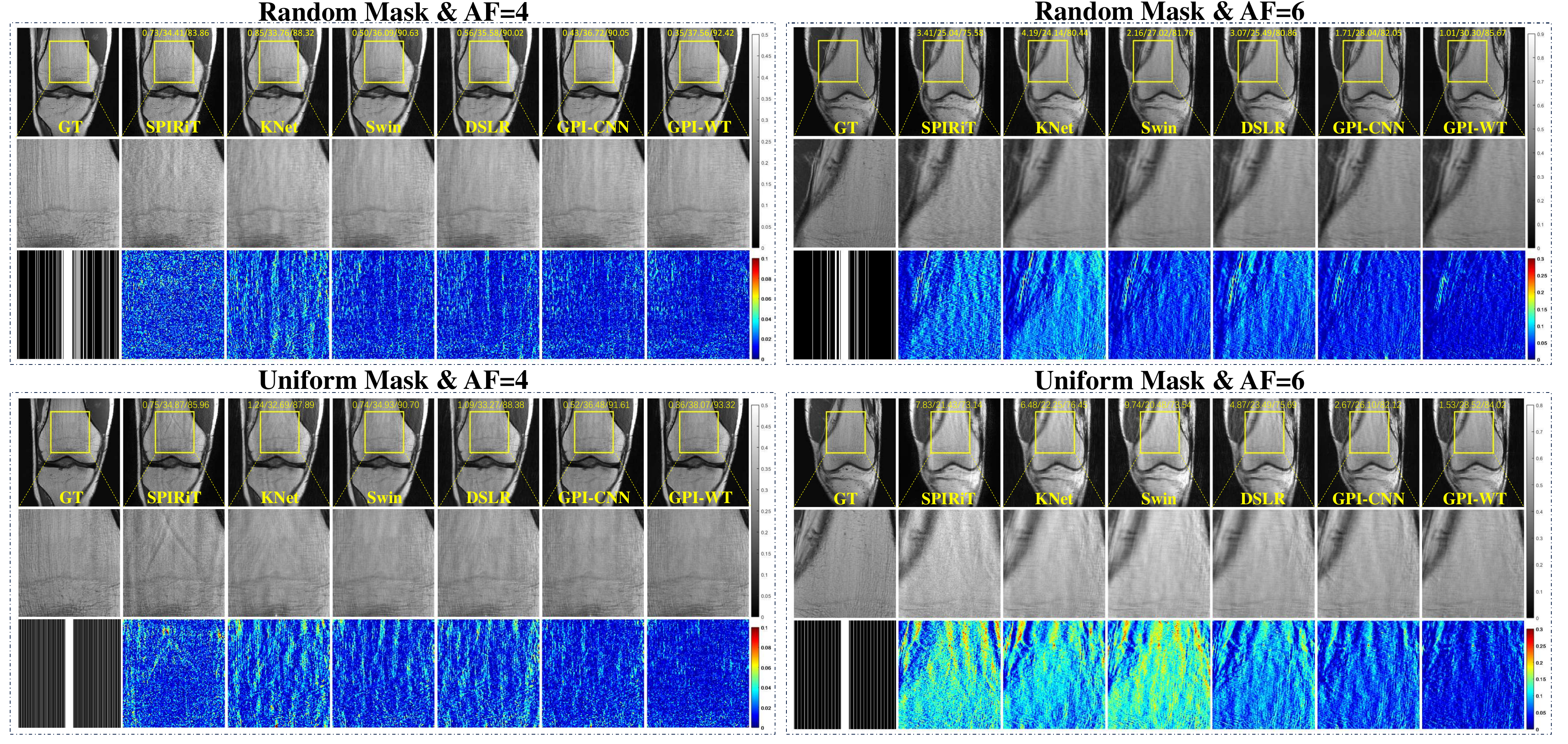}
\caption{
Reconstruction results under random and uniform masks and $\text{AF}=4,6$ with 24 ACS lines. Values of NMSE($\%$)/PSNR/SSIM($\%$) are given. The second and third rows illustrate the enlarged and error views. Grayscale bars for reconstructed images and color bars for error maps are on figures' right.
} \label{fig: knee_random}
\end{figure}

We compared our proposed approach with state-of-the-art $k$-space MR reconstruction methods, including SPIRiT~\cite{Lustig2010Spirit}, a classical training-free $k$-space interpolation method; KNet~\cite{Han2020kSpaceDL}, a DL method based on U-Net~\cite{Ronneberger2015UNet}; Swin Transformer~\cite{Liu2021SwinTransformer}; and DSLR~\cite{Pramanik2020DeepSLR}, an iterative regularization method that learns SLR constraints using CNNs. All these methods were trained exclusively in $k$-space. To further demonstrate the superiority of our GPI-WT in $k$-space, we also compared it with a CNN-based version of GPI-WT called GPI-CNN.

Comparative experiments were conducted on knee MRI data using random and uniform 2D sampling trajectories with acceleration factors (\(\text{AF}\)) of \(\text{AF}=4,6\) and an autocalibration signal (\(\text{ACS}\)) region of \(\text{ACS}=24\).    
Table \ref{tab: knee_random} summarizes the quantitative results, including average evaluation metrics. Fig. \ref{fig: knee_random} illustrates reconstructed MRI slices at various acceleration rates. Both qualitative and quantitative results demonstrate that our $k$-space GPI-WT outperforms other approaches. The zoomed-in regions in Fig. \ref{fig: knee_random} and corresponding error maps clearly demonstrate that, compared to CNN-based unrolled methods for \( k \)-space interpolation, our proposed GPI-WT significantly reduces aliasing artifacts and delivers more stable reconstructions, underscoring its enhanced capability to exploit global structural information.

\begin{table}[t]
\caption{Ablation experiment on knee dataset with mask pattern $M_{\Omega}$ and $\text{AF}=4$. $\mathcal{R}$: random mask. The optimal values are denoted in \textbf{bold} and the proposed method is highlighted with gray background..}\label{tab: knee_ab}
\centering
{\fontsize{8}{10}\selectfont 
\begin{tabular}{l|c|c|c|c}
  \hline
  Method & $M_{\Omega}$ & NMSE $(\%)\downarrow$ &PSNR $\uparrow$ & SSIM $(\%)\uparrow$ \\
  \hline  
   GPI-WT w/o LW w/o GLP &  \multirow{4}{*}{$\mathcal{R}$}   & 0.69$\pm$0.32 & 32.70$\pm$1.70 & 86.83$\pm$3.35 \\
   GPI-WT w/o GLP  &   \multirow{4}{*}{}    
   & 0.60$\pm$0.30 & 33.34$\pm$1.77 & 87.73$\pm$3.19 \\ 
   GPI-BT  &   \multirow{4}{*}{}   & 0.56$\pm$0.24 & 33.60$\pm$1.78 & 88.61$\pm$3.22 \\ 
   \rowcolor{gray!20} GPI-WT  &   \multirow{4}{*}{}   & \bfseries 0.48$\pm$0.18 & \bfseries 34.13$\pm$1.64 & \bfseries 88.94$\pm$3.24 \\ 
  \hline  
\end{tabular}}
\end{table}

\begin{figure}
\includegraphics[width=\textwidth]{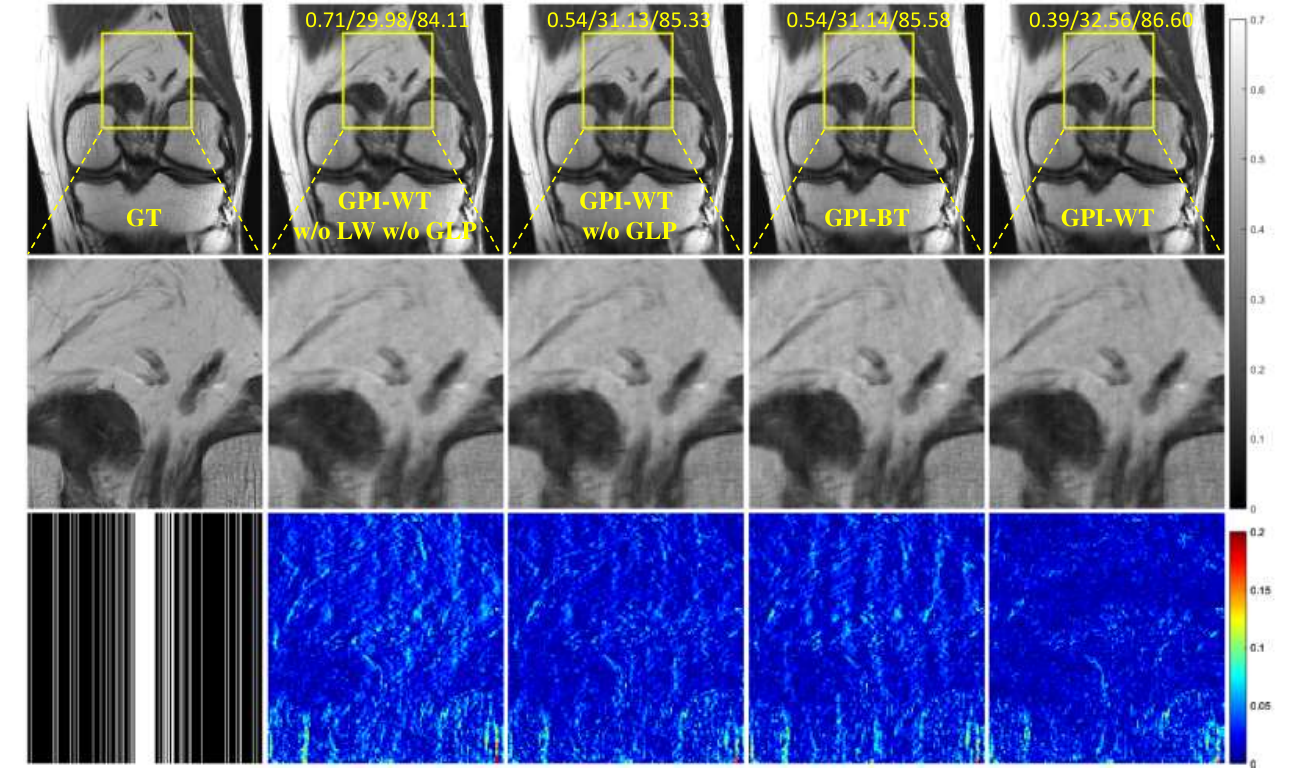}
\caption{
Ablation results under random mask at $\text{AF}=4$ with 24 ACS lines. Values of NMSE($\%$)/PSNR/SSIM($\%$) are given. The second and third rows illustrate the enlarged and error views. Grayscale bars for reconstructed images and color bars for error maps are on figure's right.
} \label{fig: knee_ab}
\end{figure}

We conducted ablation experiments using random undersampling knee data at $AF=4$ to evaluate the performance of key components in GPT-WT. Specifically, we trained and tested GPT-WT without linear window partitioning (LW) and GLP, denoted as GPT-WT w/o LW w/o GLP, which uses only square windows. To validate the effectiveness of LW, we compared it with GPT-WT without GLP but with alternating linear and square windows, denoted as GPT-WT w/o GLP. Additionally, we evaluated GPT-WT which have GLP and alternating windows against a black-box Transformer variant, denoted as GPT-BT, to assess the contributions of GLP and the white-box design. Qualitative and quantitative results are presented in Table \ref{tab: knee_ab} and Figure \ref{fig: knee_ab}.
The quantitative and qualitative ablation results clearly demonstrate the contribution of each key component in GPT-WT. Introducing LW, compared to using square windows alone, significantly improves reconstruction quality by better capturing long-range dependencies. Building upon this, the incorporation of the GLP module further enhances local structural consistency, leading to improved performance across all evaluation metrics.
To verify the advantage of the white-box design, we further compared GPT-WT with GPT-BT under the same architectural conditions. The results show that GPT-WT consistently outperforms GPT-BT, highlighting the effectiveness of interpretable priors in MRI reconstruction. As shown in Figure~\ref{fig: knee_ab}, visual comparisons confirm that GPT-WT reduces aliasing artifacts and achieves lower residual errors compared to the other variants.

\section{Conclusion}
\label{sec: conclusion}
This paper introduced GPI-WT, the first white-box Transformer for \( k \)-space interpolation, built upon the concept of GPI. Specifically, GPI was formulated from the perspective of annihilation as a novel \( k \)-space SLR model. Within this framework, the global annihilation filters were treated as learnable parameters, and the subgradients of the SLR model naturally induced a learnable attention mechanism. By unfolding the subgradient-based optimization algorithm of the SLR model into a cascaded network, the first white-box Transformer network tailored for MRI reconstruction was developed. Experimental results demonstrated that the proposed method achieved superior \( k \)-space interpolation accuracy compared to state-of-the-art approaches while offering enhanced interpretability. Currently, we focus solely on $k$-space feature learning. Future work will extend the white-box Transformer to the image domain.

\begin{credits}
\subsubsection{\ackname} This study was funded by Shenzhen Science and Technology Program under grant no. JCYJ20240813155840052; the National Key R$\&$D Program of China (2022YFA1004203, 2021YFF0501503), the National Natural Science Foundation of China (62125111, 62331028, 62476268, 62206273, 12061052, 62271474), Natural Science Fund of Inner Mongolia Autonomous Region (2024LHMS01006). 

\subsubsection{\discintname}
The authors have no competing interests to declare that are relevant to the content of this article. 
\end{credits}
\bibliographystyle{splncs04}
\bibliography{Paper-1585}
\end{document}